\documentclass{article}

    \PassOptionsToPackage{numbers, compress}{natbib}

 \usepackage[dblblindworkshop, final]{neurips_2026}
\workshoptitle{AI Foundations for Power Grids: From Models to Deployment at Scale}

\usepackage[utf8]{inputenc} 
\usepackage[T1]{fontenc}    
\usepackage{hyperref}       
\usepackage{url}            
\usepackage{booktabs}       
\usepackage{amsfonts}       
\usepackage{nicefrac}       
\usepackage{microtype}      
\usepackage{xcolor}         

\usepackage{graphicx}
\usepackage{flafter}

\usepackage{amsmath}
\usepackage{subcaption}
\usepackage{amssymb}
\usepackage{mathtools}
\usepackage{enumitem}

\title{Toward Out-of-Distribution Generalization in Neural AC Power Flow}

\author{
  Ojas Sanghi \quad Anvita Bhagavathula \quad Alvaro Carbonero \quad Priya L. Donti \\
  Department of Electrical Engineering and Computer Science \\
  Laboratory for Information and Decision Systems \\
  Massachusetts Institute of Technology, Cambridge, MA 02139 \\
  \texttt{\{osanghi, abhagava, alvarocg, donti\}@mit.edu}
}

\begin{document}

\maketitle

\begin{abstract}
  Growing uncertainty in power system operations from renewables integration, extreme weather, and data center loads challenges the reliability of traditional power flow solvers at scale. Machine learning-based proxies are a potentially promising alternative, but the prohibitive cost of generating training data for large grids makes out-of-distribution generalization a key capability for deployment; currently, models trained on small grids lack that capability. In this work, we systematically evaluate how physics-informed losses, multi-grid training, and perturbation-based data augmentation improve the zero-shot transfer of three graph neural network-based power flow models from smaller systems to a larger, unseen 500-bus system. We find that (1) physics-informed losses improve transfer relative to purely supervised objectives; (2) multi-grid training on smaller systems matches the best single grid but does not close the gap to target-grid training; (3) perturbation-based augmentation on edge attributes improves transfer by up to 54.6\%; and (4) random line/generator dropout does not consistently improve performance.
\end{abstract}

\section{Introduction}

Reliable power-system simulation is a prerequisite for operating and decarbonizing the electric grid. As intermittent
renewable energy sources, extreme weather, and energy-intensive loads increase grid uncertainty, operators must evaluate a growing
number of possible system states at operational timescales \citep{stankovski2023power, afsharinejad2021large}. In this work, we focus on AC power flow (PF), a core simulation task that determines bus voltages and power injections given loads and generator outputs (formulation in Appendix \ref{appendix-pf}). PF calculations underpin workflows such as topology optimization 
and contingency analysis, 
and serve as a feasibility constraint in the closely related AC optimal power flow (ACOPF) problem
\citep{gomez2018electric}.

Traditional PF solvers rely on iterative methods such as Newton-Raphson (NR) and struggle to tractably scale to solve the large number of scenarios required in real time \citep{d2021comparing}. Graph neural network (GNN)-based surrogates have been explored as promising alternatives, offering substantial speedups on large systems~\citep{hansen2022power}, but face several barriers to deployment. One such barrier is their reliance on highly diverse and realistic training data, whose creation is computationally expensive for realistic grid sizes \citep{birchfield2016grid}. The PF$\Delta$ benchmark \citep{rivera2026pf} shows that models trained on a single small or medium-scale grid generalize poorly to larger systems, despite strong in-distribution performance. However, data diversity does improve performance within a fixed grid size: training on $N-1$ perturbations improves performance on more challenging $N-2$ perturbations.

\begin{figure}[t]
    \centering
    \includegraphics[width=0.90\linewidth]{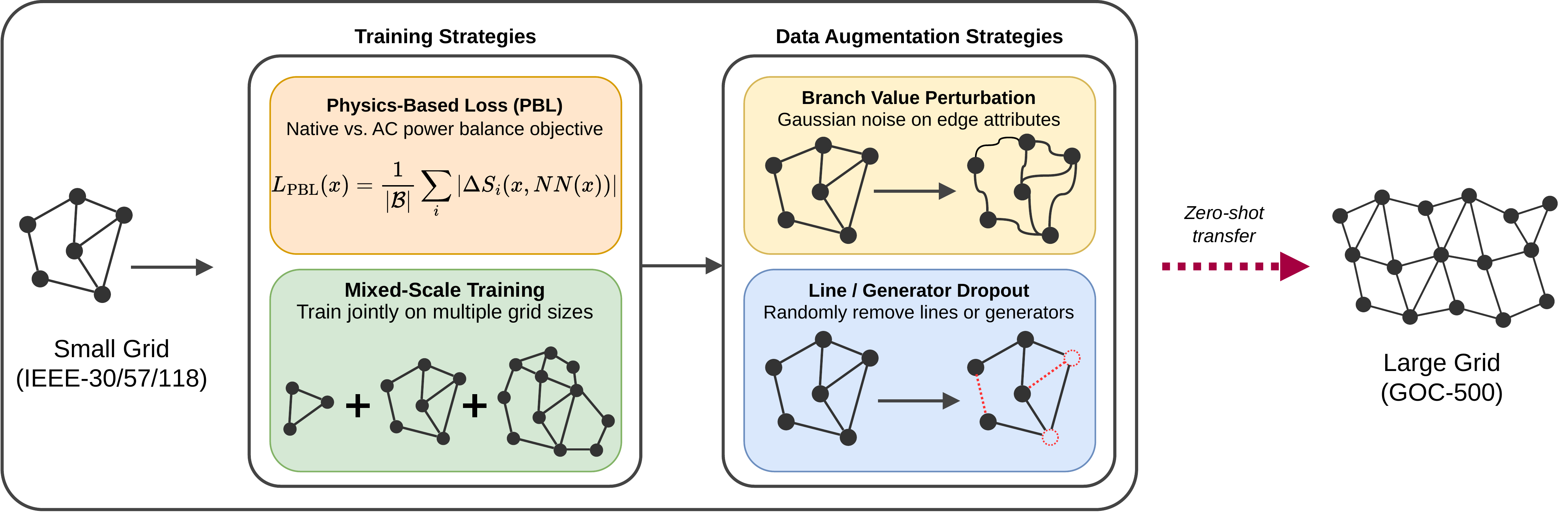}
    \caption{Overview of our evaluation framework. We train GNN-based power flow approximators on small benchmark grids (IEEE-30/57/118) with a physics-based loss (Power Balance Loss, or PBL) and mixed-scale training, apply branch value perturbation and line/generator dropout, then evaluate zero-shot transfer to the larger, out-of-distribution GOC-500 grid.}
    \vspace{-10pt}
    \label{fig:evaluation-overview}
\end{figure}

These findings suggest that training strategy may be as important as model architecture for generalization, yet it remains unclear which practical interventions improve zero-shot transfer to larger unseen grids. In this work, we systematically evaluate three levers: (1) physics-informed objectives, (2) multi-grid training for scale extrapolation, and (3) physically motivated data augmentation. Concretely, we: (1) empirically study two training strategies and two data augmentation strategies across three GNN-based power flow models (\textit{CANOS-PF}, \textit{PowerFlowNet}, and \textit{GraphConv}) trained on smaller IEEE systems and evaluated zero-shot on GOC-500 (Figure \ref{fig:evaluation-overview}); (2) establish that physics-informed losses consistently improve zero-shot performance on GOC-500 relative to native supervised objectives; (3) show that branch-value perturbation improves transfer for edge-aware GNN architectures by up to 54.6\%; and (4) find that multi-grid training on smaller systems (IEEE-30/57/118) and line/generator dropout do not close the gap to training directly on the target grid.

\section{Related work}

\textbf{Surrogate Modeling for Power System Simulation.} A growing body of work applies machine learning to accelerate power system optimization and simulation tasks such as ACOPF and PF. Feedforward networks trained on fixed topologies have been used to approximate ACOPF solutions \citep{donti2021dc3, nguyen2026fsnet, pan2022deepopf}. GNNs that better capture grid structure and topology changes via interaction-network, graph transformer, and foundation-model approaches have been used to solve ACOPF, PF, and reconfiguration tasks \citep{piloto2024canos, ghamizi2024opf, mahto2024gat, wu2025universalgraphlearningpower, HAMANN20243245}. In parallel, recent efforts have focused on large-scale data generation and systematic benchmarking \citep{lovett2024opfdatalargescaledatasetsac, rivera2026pf, Joswig_Jones_2022}, enabling evaluation across operating conditions and grid sizes  and revealing current limitations in out-of-distribution transfer. We build on these benchmarking efforts, testing whether established training interventions can close the transfer gap.

\textbf{Generalizability of GNNs.} More broadly, GNN generalizability has been studied from several perspectives, including regularization \citep{buffelli2022sizeshiftregregularizationmethodimproving}, size-generalization strategies \citep{tang2023understandinggeneralizationgraphneural}, spectral analyses \citep{li2025tacklingsizegeneralizationgraph}, dropout-based stochasticity \citep{10.5555/2627435.2670313, rong2020dropedgedeepgraphconvolutional}, and pretrained graph foundation models \citep{HAMANN20243245, galkin2024foundationmodelsknowledgegraph}. Despite this progress, size generalization remains an open challenge for GNNs; models trained on small graphs degrade under scale extrapolation and topology shift across application domains. Power flow compounds the problem, since predictions must additionally satisfy nonlinear physical constraints whose conditioning changes with network size. Here, we test the impact of two methods from this literature: diversifying the training distribution and introducing stochastic structural dropout.

\section{Methods}

\begin{figure}[t]
  \centering
   \includegraphics[width=\linewidth]{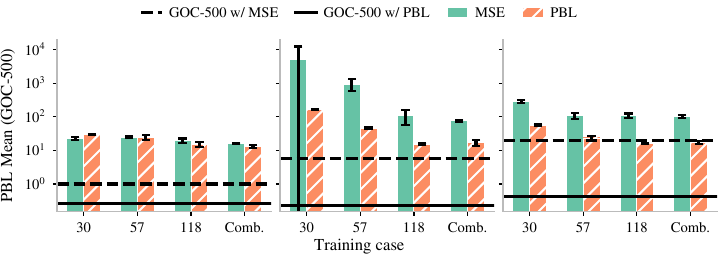}
  \\
  \begin{subfigure}[t]{0.32\linewidth}\vspace{-5pt}\caption{CANOS-PF}\label{fig:mg-canos}\end{subfigure}
  \begin{subfigure}[t]{0.32\linewidth}\vspace{-5pt}\caption{PowerFlowNet}\label{fig:mg-pfnet}\end{subfigure}
  \begin{subfigure}[t]{0.32\linewidth}\vspace{-5pt}\caption{GraphConv}\label{fig:mg-gcn}\end{subfigure}
  \caption{Effect of loss function and multi-grid training on GOC-500 performance
           across all three GNNs. Horizontal lines show the same architecture trained directly on GOC-500 with MSE or PBL, respectively, as an in-distribution reference. Error bars = standard deviation
           over 3 training runs.}
        \vspace{-4pt}
    \label{fig:multigrid-all}
\end{figure}

We train three models (\textit{CANOS-PF}, \textit{PowerFlowNet}, and \textit{GraphConv}) with two training strategies and two data augmentation methods, and measure zero-shot performance on PF$\Delta$'s GOC-500 test set. PF$\Delta$ consists of 859,800 solved AC power flow instances across IEEE-14/30/57/118 and GOC-500/2000 \citep{birchfield2016grid, babaeinejadsarookolaee2021Powerc}. Models are evaluated on how well solutions respect power balance via mean Power Balance Loss (PBL), defined in PF$\Delta$ as the average mismatch $|\big(\Delta S(x, \hat{y})\big)_i|$ across all buses $i$, where $x$ is an input configuration, $\hat{y}_{\text{model}} = \mathrm{NN}(x)$ is a model's output, and $\Delta S = \operatorname{diag}(v)\bar{Y}\bar{v} - \left[(p_g - p_d) + (q_g - q_d)j\right]$ is the power balance mismatch.

\textbf{Trained Models.} \textbf{CANOS-PF} \citep{rivera2026pf} is an adaptation of the AC-OPF solver CANOS \citep{piloto2024canos} that leverages interaction networks and data heterogeneity through specialized encoders and decoders. It also analytically derives the real and reactive power generation at the slack bus. \textbf{PowerFlowNet} \citep{lin2024powerflownet} combines bus-typed embeddings of the input features and mask embeddings with TAGConv message passing layers. \textbf{GraphConv} \citep{morris2018weisfeiler} is a simple message passing architecture that does not use edge attributes, in contrast to the other two PF$\Delta$-benchmarked models described above.

\textbf{Training Strategies.} We first vary the loss function of the model and the set of grids it is trained on.
\begin{itemize}[leftmargin=*]
    \item \textbf{Loss Functions.} We compare each model trained on its native loss against training directly on PBL. With input $x$, solver solution $y$, model output $\hat{y}_{\text{model}} = \mathrm{NN}(x)$, the native losses are $\mathrm{MSE}(\hat{y}, y)$ for PowerFlowNet and GraphConv, and $\mathrm{MSE}(\hat{y}, y) + \lambda C(x, y, \hat{y})$ for CANOS-PF, where $\lambda$ is a hyperparameter and $C$ penalizes power balance mismatch and branch flow deviation from ground truth \citep{rivera2026pf}. Training on PBL instead minimizes, over a network with buses $\mathcal{B}$, the loss $L_{\text{PBL}}(x) := \frac{1}{|\mathcal{B}|}\sum_{i \in \mathcal{B}}|\Delta S_i(x, \mathrm{NN}(x))|$. By directly penalizing power balance mismatch rather than deviation from a solver's solution, PBL incentivizes the model to behave as a solver itself, potentially improving OOD robustness; being self-supervised, it requires no precomputed solutions, enabling data augmentation.
    
    \item \textbf{Multi-Grid Training.} Given the intuition that topological diversity during training improves generalization to unseen topologies, we train jointly on IEEE-30, IEEE-57, and IEEE-118, evaluating on GOC-500. We compare to training
    directly on GOC-500 and individually on each smaller grid.
\end{itemize}

\begin{figure}[t]
    \centering
    \includegraphics[width=\linewidth]{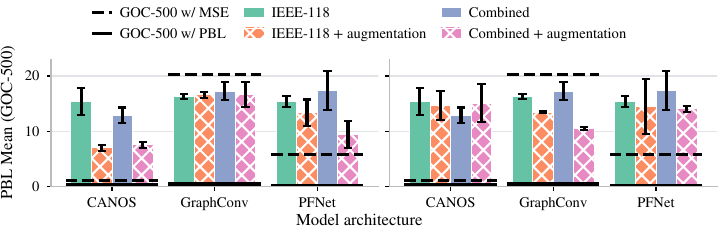}
    \\[-0.3em]
    \begin{subfigure}[t]{0.49\linewidth}
        \centering \vspace{-5pt}
        \subcaption{Branch perturbation}
        \label{fig:branch_perturbation}
    \end{subfigure}
    \hfill
    \begin{subfigure}[t]{0.49\linewidth}
        \centering
        \vspace{-5pt}\subcaption{Line/generator N-2 dropout}
        \label{fig:line_gen_dropout}
    \end{subfigure}
    \caption{Effect of 
    data augmentation strategies on GOC-500 performance under PBL training. ``Combined'' denotes joint training on IEEE-30/%
    57/%
    118. Horizontal lines are as defined in Figure~\ref{fig:multigrid-all}. 
    }
    \label{fig:augmentation-all}
    \vspace{-6pt}
\end{figure}

\textbf{Data Augmentation.} We implement two data augmentation strategies to increase the diversity of the training distribution, applied under PBL training to the strongest-performing settings on GOC-500: IEEE-118 and the combined IEEE-30/57/118 dataset.
\begin{itemize}[leftmargin=*]
    \item \textbf{Line/Generator Dropout.} Every training set already includes 0 to 2 contingencies; we add 2 additional random contingencies (2 to 4 total), testing whether magnified topological diversity prevents overfitting to training topologies. Each added contingency removes a randomly selected line or generator with equal probability; a PV bus left without a generator becomes a PQ bus. Augmented samples are not tested for connectivity or feasibility.
    
    \item \textbf{Branch Value Perturbation.} We perturb branch parameters with Gaussian noise, addressing the lack of branch value variation in PF$\Delta$, which risks overfitting to narrow numerical ranges. For each branch $e$ we draw $y_e \sim \mathcal{N}(0,1)$; each branch value $x$ for the series impedance and shunt admittance terms becomes $f_\sigma(x) := x \cdot \text{ReLU}(1 + \sigma y_e)$, where $\sigma \in \mathbb{R}$ is a hyperparameter per model.\footnote{See Appendix~\ref{appendix-hyperparameter} for hyperparameter tuning details and the selected $\sigma$ values per model.} The same $y_e$ is used for all of a branch's perturbed values, and ReLU ensures non-negativity.
\end{itemize}

\section{Results and discussion}
\label{sec:experiments}

\textbf{Physics-Informed Training Loss.} PBL training substantially improves OOD performance on GOC-500 for the purely supervised models, PowerFlowNet and GraphConv
across all training-grid settings (Fig.~\ref{fig:multigrid-all}).
However, this is not the case for CANOS-PF, 
suggesting CANOS-PF's native 
loss provides sufficient physics-aware regularization.

\textbf{Multi-Grid Training.} 
Combined IEEE-30/57/118 training is generally on par with the best single-grid setting (IEEE-118), but does not close the performance gap with models trained directly on GOC-500 with PBL (Figure~\ref{fig:multigrid-all}).  Diversity across smaller topologies alone is insufficient for cross-grid transfer, suggesting a need for limited use of target-grid information or stronger global representations.

\textbf{Branch Perturbation.} Under PBL training, branch perturbation improves CANOS-PF's zero-shot transfer to GOC-500 by 54.6\% when trained on IEEE-118 and by 41.6\% when trained jointly on IEEE-30/57/118, relative to unaugmented training (Figure~\ref{fig:branch_perturbation}). 
PowerFlowNet also benefits, particularly in the combined-grid setting. GraphConv, which lacks edge attributes, shows no significant improvement, suggesting that branch perturbations benefit models that exploit heterogeneous edge interactions.

\textbf{Line/Generator Dropout.} Random line/generator dropout does not consistently improve transfer (Figure~\ref{fig:line_gen_dropout}). While it improves GraphConv's performance, it can degrade performance for CANOS-PF and PowerFlowNet relative to the unaugmented baseline. This inconsistency may be caused by infeasible instances, since augmented samples are not tested for connectivity or feasibility, although prior self-supervised work trains successfully on data that includes infeasible instances \citep{nguyen2026scalable}. We leave further analysis of this augmentation strategy to future work. These results suggest that physically feasible variation in edge attributes (i.e., branch perturbations) may be more effective for zero-shot transfer than random outages, which can produce unrealistic or infeasible configurations.

\textbf{Worst-case behaviour.} We repeat the analysis using the maximum per-bus mismatch (Appendix~\ref{appendix-worstcase}). Every conclusion is unchanged; PBL training and branch perturbation remain effective interventions.

\section{Conclusion and further work}
We studied whether practical training interventions can improve zero-shot generalization of GNN-based AC power flow models to an unseen 500-bus system. We find that physics-informed training (PBL) is the most reliable intervention, consistently reducing zero-shot OOD residuals across architectures; branch-value perturbation improves transfer for edge-aware architectures; multi-grid training does not close the gap to target-grid training; and random line/generator dropout is inconsistent and can degrade performance. Our findings suggest that effective training strategies and data augmentation should be grounded in physically meaningful variation.

Closing this gap -- a prerequisite for trustworthy deployment -- will require jointly advancing training and architecture through scale-transfer inductive biases (e.g., scale-invariant representations); physics-constrained augmentation co-designed with edge-aware architectures; few-shot fine-tuning on limited target-grid data as a middle ground between full retraining and zero-shot deployment; and evaluation across more topologically diverse systems beyond GOC-500.

\begin{ack}

This material is based upon work supported by the National Science Foundation Graduate Research Fellowship Program under Grant No. 2141064; the U.S. Department of Energy, Office of Science, Office of Advanced Scientific Computing Research, Department of Energy Computational Science Graduate Fellowship under Award Number DE-SC0025528; and the U.S. National Science Foundation (award \#2325956). Any opinions, findings, and conclusions or recommendations expressed in this material are those of the authors and do not necessarily reflect the views of the National Science Foundation. The authors acknowledge the MIT Office of Research Computing and Data for providing high performance computing resources that have contributed to the research results reported within this paper.

\end{ack}

\bibliography{bibliography}

@article{rivera2026pf,
  title={{PF}{$\Delta$}: A Benchmark Dataset for Power Flow under Load, Generation, and Topology Variations},
  author={Rivera Him, Ana and Bhagavathula, Anvita and Carbonero, Alvaro and Donti, Priya},
  journal={Advances in Neural Information Processing Systems},
  volume={38},
  year={2025}
}

@inproceedings{donti2021dc3,
  title={{DC3}: A Learning Method for Optimization with Hard Constraints},
  author={Donti, Priya L. and Rolnick, David and Kolter, J. Zico},
  booktitle={International Conference on Learning Representations},
  year={2021}
}

@article{nguyen2026fsnet,
  title={{FSNet}: Feasibility-Seeking Neural Network for Constrained Optimization with Guarantees},
  author={Nguyen, Hoang T. and Donti, Priya L.},
  journal={Advances in Neural Information Processing Systems},
  volume={38},
  pages={39670--39708},
  year={2025}
}

@article{pan2022deepopf,
  title={{DeepOPF}: A Feasibility-Optimized Deep Neural Network Approach for {AC} Optimal Power Flow Problems},
  author={Pan, Xiang and Chen, Minghua and Zhao, Tianyu and Low, Steven H.},
  journal={IEEE Systems Journal},
  volume={17},
  number={1},
  pages={673--683},
  year={2022},
  publisher={IEEE}
}

@article{lin2024powerflownet,
  title={{PowerFlowNet}: Power Flow Approximation Using Message Passing Graph Neural Networks},
  author={Lin, Nan and Orfanoudakis, Stavros and Cardenas, Nathan Ordonez and Giraldo, Juan S. and Vergara, Pedro P.},
  journal={International Journal of Electrical Power \& Energy Systems},
  volume={160},
  pages={110112},
  year={2024},
  publisher={Elsevier}
}

@article{piloto2024canos,
  title={{CANOS}: A Fast and Scalable Neural {AC-OPF} Solver Robust to {N-1} Perturbations},
  author={Piloto, Luis and Liguori, Sofia and Madjiheurem, Sephora and Zgubic, Miha and Lovett, Sean and Tomlinson, Hamish and Elster, Sophie and Apps, Chris and Witherspoon, Sims},
  journal={arXiv preprint arXiv:2403.17660},
  year={2024}
}

@inproceedings{ghamizi2024opf,
  title={{OPF-HGNN}: Generalizable Heterogeneous Graph Neural Networks for {AC} Optimal Power Flow},
  author={Ghamizi, Salah and Ma, Aoxiang and Cao, Jun and Cortes, Pedro Rodriguez},
  booktitle={2024 IEEE Power \& Energy Society General Meeting (PESGM)},
  pages={1--5},
  year={2024},
  organization={IEEE}
}

@article{mahto2024gat,
  title={{GAT-ADNet}: Leveraging Graph Attention Network for Optimal Power Flow in Active Distribution Network with High Renewables},
  author={Mahto, Dinesh Kumar and Bukya, Mahipal and Kumar, Rajesh and Mathur, Akhilesh and Saini, Vikash Kumar},
  journal={IEEE Access},
  volume={12},
  pages={185728--185739},
  year={2024},
  publisher={IEEE}
}

@article{stankovski2023power,
  title={Power Blackouts in {Europe}: Analyses, Key Insights, and Recommendations from Empirical Evidence},
  author={Stankovski, Andrej and Gjorgiev, Blazhe and Locher, Leon and Sansavini, Giovanni},
  journal={Joule},
  volume={7},
  number={11},
  pages={2468--2484},
  year={2023},
  publisher={Elsevier}
}

@article{afsharinejad2021large,
  title={Large-Scale Data Analytics for Resilient Recovery Services from Power Failures},
  author={Afsharinejad, A. H. and Ji, C. and Wilcox, R.},
  journal={Joule},
  volume={5},
  pages={2504--2520},
  year={2021}
}

@book{gomez2018electric,
  title={Electric Energy Systems: Analysis and Operation},
  author={G{\'o}mez-Exp{\'o}sito, Antonio and Conejo, Antonio J. and Ca{\~n}izares, Claudio},
  year={2018},
  publisher={CRC Press}
}

@misc{wu2025universalgraphlearningpower,
  title={Universal Graph Learning for Power System Reconfigurations: Transfer Across Topology Variations},
  author={Tong Wu and Anna Scaglione and Sandy Miguel and Daniel Arnold},
  year={2025},
  eprint={2509.08672},
  archivePrefix={arXiv},
  primaryClass={eess.SY},
  url={https://arxiv.org/abs/2509.08672}
}

@article{HAMANN20243245,
  title = {Foundation Models for the Electric Power Grid},
  journal = {Joule},
  volume = {8},
  number = {12},
  pages = {3245--3258},
  year = {2024},
  issn = {2542-4351},
  doi = {10.1016/j.joule.2024.11.002},
  url = {https://www.sciencedirect.com/science/article/pii/S2542435124004707},
  author = {Hendrik F. Hamann and Blazhe Gjorgiev and Thomas Brunschwiler and Leonardo S.A. Martins and Alban Puech and Anna Varbella and Jonas Weiss and Juan Bernabe-Moreno and Alexandre Blondin Massé and Seong Lok Choi and Ian Foster and Bri-Mathias Hodge and Rishabh Jain and Kibaek Kim and Vincent Mai and François Mirallès and Martin {De Montigny} and Octavio Ramos-Leaños and Hussein Suprême and Le Xie and El-Nasser S. Youssef and Arnaud Zinflou and Alexander Belyi and Ricardo J. Bessa and Bishnu Prasad Bhattarai and Johannes Schmude and Stanislav Sobolevsky}
}

@misc{lovett2024opfdatalargescaledatasetsac,
  title={{OPFData}: Large-Scale Datasets for {AC} Optimal Power Flow with Topological Perturbations},
  author={Sean Lovett and Miha Zgubic and Sofia Liguori and Sephora Madjiheurem and Hamish Tomlinson and Sophie Elster and Chris Apps and Sims Witherspoon and Luis Piloto},
  year={2024},
  eprint={2406.07234},
  archivePrefix={arXiv},
  primaryClass={cs.LG},
  url={https://arxiv.org/abs/2406.07234}
}

@inproceedings{Joswig_Jones_2022,
  title={{OPF-Learn}: An Open-Source Framework for Creating Representative {AC} Optimal Power Flow Datasets},
  author={Joswig-Jones, Trager and Baker, Kyri and Zamzam, Ahmed S.},
  booktitle={2022 IEEE Power \& Energy Society Innovative Smart Grid Technologies Conference (ISGT)},
  pages={1--5},
  year={2022},
  month=apr,
  publisher={IEEE},
  doi={10.1109/ISGT50606.2022.9817509},
  url={http://dx.doi.org/10.1109/ISGT50606.2022.9817509}
}

@article{hansen2022power,
  title={Power Flow Balancing with Decentralized Graph Neural Networks},
  author={Hansen, Jonas Berg and Anfinsen, Stian Normann and Bianchi, Filippo Maria},
  journal={IEEE Transactions on Power Systems},
  volume={38},
  number={3},
  pages={2423--2433},
  year={2022},
  publisher={IEEE}
}

@article{d2021comparing,
  title={Comparing Different Approaches for Solving Large Scale Power-Flow Problems with the {Newton-Raphson} Method},
  author={{D'Orto}, Manolo and Sj{\"o}blom, Svante and Chien, Lung Sheng and Axner, Lilit and Gong, Jing},
  journal={IEEE Access},
  volume={9},
  pages={56604--56615},
  year={2021},
  publisher={IEEE}
}

@misc{buffelli2022sizeshiftregregularizationmethodimproving,
  title={{SizeShiftReg}: A Regularization Method for Improving Size-Generalization in Graph Neural Networks},
  author={Davide Buffelli and Pietro Liò and Fabio Vandin},
  year={2022},
  eprint={2207.07888},
  archivePrefix={arXiv},
  primaryClass={cs.LG},
  url={https://arxiv.org/abs/2207.07888}
}

@misc{tang2023understandinggeneralizationgraphneural,
  title={Towards Understanding the Generalization of Graph Neural Networks},
  author={Huayi Tang and Yong Liu},
  year={2023},
  eprint={2305.08048},
  archivePrefix={arXiv},
  primaryClass={cs.LG},
  url={https://arxiv.org/abs/2305.08048}
}

@misc{li2025tacklingsizegeneralizationgraph,
  title={Tackling Size Generalization of Graph Neural Networks on Biological Data from a Spectral Perspective},
  author={Gaotang Li and Danai Koutra and Yujun Yan},
  year={2025},
  eprint={2305.15611},
  archivePrefix={arXiv},
  primaryClass={cs.LG},
  url={https://arxiv.org/abs/2305.15611}
}

@article{birchfield2016grid,
  title={Grid Structural Characteristics as Validation Criteria for Synthetic Networks},
  author={Birchfield, Adam B. and Xu, Ti and Gegner, Kathleen M. and Shetye, Komal S. and Overbye, Thomas J.},
  journal={IEEE Transactions on Power Systems},
  volume={32},
  number={4},
  pages={3258--3265},
  year={2016},
  publisher={IEEE}
}

@article{10.5555/2627435.2670313,
  author = {Srivastava, Nitish and Hinton, Geoffrey and Krizhevsky, Alex and Sutskever, Ilya and Salakhutdinov, Ruslan},
  title = {Dropout: A Simple Way to Prevent Neural Networks from Overfitting},
  journal = {Journal of Machine Learning Research},
  volume = {15},
  number = {1},
  pages = {1929--1958},
  year = {2014}
}

@misc{rong2020dropedgedeepgraphconvolutional,
  title={{DropEdge}: Towards Deep Graph Convolutional Networks on Node Classification},
  author={Yu Rong and Wenbing Huang and Tingyang Xu and Junzhou Huang},
  year={2020},
  eprint={1907.10903},
  archivePrefix={arXiv},
  primaryClass={cs.LG},
  url={https://arxiv.org/abs/1907.10903}
}

@misc{galkin2024foundationmodelsknowledgegraph,
  title={Towards Foundation Models for Knowledge Graph Reasoning},
  author={Mikhail Galkin and Xinyu Yuan and Hesham Mostafa and Jian Tang and Zhaocheng Zhu},
  year={2024},
  eprint={2310.04562},
  archivePrefix={arXiv},
  primaryClass={cs.CL},
  url={https://arxiv.org/abs/2310.04562}
}

@misc{babaeinejadsarookolaee2021Powerc,
  title={The {Power Grid Library} for Benchmarking {AC} Optimal Power Flow Algorithms},
  author={Sogol Babaeinejadsarookolaee and Adam Birchfield and Richard D. Christie and Carleton Coffrin and Christopher DeMarco and Ruisheng Diao and Michael Ferris and Stephane Fliscounakis and Scott Greene and Renke Huang and Cedric Josz and Roman Korab and Bernard Lesieutre and Jean Maeght and Terrence W. K. Mak and Daniel K. Molzahn and Thomas J. Overbye and Patrick Panciatici and Byungkwon Park and Jonathan Snodgrass and Ahmad Tbaileh and Pascal Van Hentenryck and Ray Zimmerman},
  year={2021},
  eprint={1908.02788},
  archivePrefix={arXiv},
  primaryClass={math.OC},
  url={https://arxiv.org/abs/1908.02788}
}

@misc{morris2018weisfeiler,
  title = {{Weisfeiler} and {Leman} Go Neural: Higher-Order Graph Neural Networks},
  author = {Christopher Morris and Martin Ritzert and Matthias Fey and William L. Hamilton and Jan Eric Lenssen and Gaurav Rattan and Martin Grohe},
  year = {2018},
  eprint = {1810.02244},
  archivePrefix = {arXiv},
  primaryClass = {cs.LG},
  url = {https://arxiv.org/abs/1810.02244}
}

@article{nguyen2026scalable,
  title = {Scalable Self-Supervised Learning for Multiphase {AC-OPF} in Distribution Systems with Topology Reconfiguration},
  author = {Nguyen, Hoang T. and Liu, Shaohui and Sen Biswas, Reetam and Pendyala, Varsha and Virani, Nurali and Deka, Deepjyoti and Donti, Priya L.},
  journal = {arXiv preprint arXiv:2608.25095},
  year = {2026}
}
\bibliographystyle{unsrtnat}

\newpage
\appendix

\section{AC power flow: problem formulation}
\label{appendix-pf}

This appendix summarizes the AC power flow formulation used throughout this work. Our notation and high-level presentation follow the formulation in PF$\Delta$ \citep{rivera2026pf}; we restate the relevant equations here for completeness.

A power network is represented as a graph with buses $\mathcal{B}$ and transmission lines $\mathcal{L} \subseteq \{(i,j) \mid i,j \in \mathcal{B},\; i \neq j\}$. At each bus $i \in \mathcal{B}$, the AC power flow state is described by four quantities: voltage magnitude $|v_i|$, voltage angle $\theta_i$, net active power injection $p_{\text{net},i}$, and net reactive power injection $q_{\text{net},i}$. Different bus types specify different pairs of known and unknown quantities. Table \ref{tab:bus-types-appendix} summarizes the standard slack, PQ, and PV bus conventions used in this work.

\begin{table}[h]
\caption{Known and unknown quantities for each bus type in AC power flow.}
\label{tab:bus-types-appendix}
\centering
\begin{tabular}{lll}
\toprule
\textbf{Bus Type} & \textbf{Known (input) quantities} & \textbf{Unknown (output) quantities} \\
\midrule
Slack          & $|v|, \theta$                    & $p_{\text{net}}, q_{\text{net}}$ \\
Load (PQ)      & $p_{\text{net}}, q_{\text{net}}$ & $|v|, \theta$                    \\
Generator (PV) & $p_{\text{net}}, |v|$            & $\theta, q_{\text{net}}$         \\
\bottomrule
\end{tabular}
\end{table}

Given the known quantities in Table \ref{tab:bus-types-appendix}, the power flow problem solves for the remaining bus variables such that active and reactive power are balanced at every bus. Let $\theta_{ij} := \theta_i - \theta_j$, and let $G_{ij}$ and $B_{ij}$ denote the conductance and susceptance entries of the network admittance matrix. The active and reactive power balance residuals are

\begin{equation}
    0 = \Delta p_i := p_{\text{net},i} - |v_i| \sum_{j=1}^{n} |v_j| 
    \left( G_{ij} \cos \theta_{ij} + B_{ij} \sin \theta_{ij} \right)
    \quad \forall i \in \mathcal{B},
    \label{eq:active-power-mismatch-app}
\end{equation}

\begin{equation}
    0 = \Delta q_i := q_{\text{net},i} - |v_i| \sum_{j=1}^{n} |v_j| 
    \left( G_{ij} \sin \theta_{ij} - B_{ij} \cos \theta_{ij} \right)
    \quad \forall i \in \mathcal{B}.
    \label{eq:reactive-power-mismatch-app}
\end{equation}

Once the bus voltages are known, active and reactive line flows can be computed for each line $(i,j) \in \mathcal{L}$ as

\begin{equation}
    p_{ij} = |v_i||v_j|
    \left(G_{ij} \cos \theta_{ij} + B_{ij} \sin \theta_{ij}\right) 
    - G_{ij}|v_i|^2
    \quad \forall (i,j) \in \mathcal{L},
    \label{eq:active-line-flow-app}
\end{equation}

\begin{equation}
    q_{ij} = |v_i||v_j|
    \left(G_{ij} \sin \theta_{ij} - B_{ij} \cos \theta_{ij}\right) 
    + |v_i|^2\left(B_{ij} - b_{s,ij}\right)
    \quad \forall (i,j) \in \mathcal{L},
    \label{eq:reactive-line-flow-app}
\end{equation}

where $b_{s,ij}$ is the shunt susceptance of line $(i,j)$. Equivalently, the nodal power balance equations can be written in compact complex form as

\begin{equation}
    \operatorname{diag}(v)\,\bar{Y}\bar{v} = (p_g - p_d) + (q_g - q_d)j,
    \label{eq:power-flow-compact-app}
\end{equation}

where $p_g$ and $q_g$ are generator active and reactive injections, $p_d$ and $q_d$ are active and reactive demands, $Y$ is the network admittance matrix, and $\bar{(\cdot)}$ denotes complex conjugation.

Classical AC power flow solvers typically solve first for the unknown voltage quantities. In particular, Newton-Raphson iteratively solves the nonlinear balance equations for voltage magnitudes and angles at PQ buses and voltage angles at PV buses. The remaining quantities, including slack-bus injections, generator reactive power outputs, and line flows, are then computed from the solved voltage state.

\newpage
\section{Power balance loss}
\label{appendix-pbl}

When training a machine learning model to solve the power flow equations, the model inputs $x$ encode the known quantities at each bus (per bus type, per Table~\ref{tab:bus-types-appendix}) together with branch parameters for each line. The model outputs $\hat{y} = \mathrm{NN}(x)$ predict the unknown bus quantities. 

Power Balance Loss (PBL) is then computed directly from the predicted solution without requiring a precomputed ground-truth label, as the magnitude of the complex power mismatch at each bus:

\begin{equation}
    |\Delta S_i| = \sqrt{(\Delta p_i)^2 + (\Delta q_i)^2} \quad \forall i \in \mathcal{B},
\end{equation}
and mean PBL is averaged across all buses:
\begin{equation}
    L_{\text{PBL}}(x) = \frac{1}{|\mathcal{B}|}\sum_{i \in \mathcal{B}} |\Delta S_i(x,\, \mathrm{NN}(x))|.
\end{equation}

\section{Hyperparameter tuning}
\label{appendix-hyperparameter}

We tuned each model's hyperparameters using a grid search strategy, selecting final configurations based on validation performance using each model's native training loss. Validation sets were fixed random subsamples of 10\% of the training dataset, consistent with the setup in PF$\Delta$ \citep{rivera2026pf}. We employed an early stopping strategy that halts training if the epoch with the best validation PBL (Mean) occurred more than 15 epochs prior (or 30 epochs for CANOS-PF), or if the validation error has not changed by more than 1\% for 10 consecutive epochs (or 20 for CANOS-PF). All models used a batch size of 64 and the Adam optimizer with no weight decay. For each model, for branch perturbation experiments, we swept $\sigma \in \{0.01, 0.03, 0.10, 0.30, 0.50\}$ on IEEE-118 under PBL training and selected the value minimizing IEEE-118 mean PBL. We did not tune hyperparameters on GOC-500; the models trained directly on GOC-500 reuse the configurations below, which might lead to an underestimation of the generalization gap.

Model-specific details are as follows:

\begin{itemize}

    \item \textbf{CANOS-PF}: We performed a grid search over \texttt{hidden\_dim}, \texttt{k\_steps} (the depth of the message-passing network), and the learning rate. The model was trained for 200 epochs with a \texttt{SequentialLR} schedule: a linear warmup from a factor of 0.01 to 1.0 over the first 5 epochs, followed by a \texttt{StepLR} decay with $\gamma = 0.9$ every 4 epochs. The finalized parameters were \texttt{hidden\_dim} = 128, \texttt{k\_steps} = 20, and learning rate = 0.0005. For branch perturbation experiments, the best $\sigma$ on IEEE-118 was $\sigma = 0.10$.

    \item \textbf{PowerFlowNet}: We conducted a grid search over learning rate $\in \{10^{-4}, 5{\times}10^{-4}, 10^{-3}, 5{\times}10^{-3}\}$, with all other architectural hyperparameters fixed following the original implementation \citep{lin2024powerflownet}. The model was trained for 150 epochs with no learning rate schedule. The final configuration used \texttt{hidden\_dim} = 256, \texttt{n\_gnn\_layers} = 4, \texttt{K} = 3 (TAGConv receptive field), and a dropout rate of 0.2, with a learning rate of $10^{-4}$. For branch perturbation experiments, the best $\sigma$ on IEEE-118 was $\sigma = 0.03$.

    \item \textbf{GraphConv}: We conducted a grid search over learning rate $\in \{10^{-4}, 5{\times}10^{-4}, 10^{-3}, 5{\times}10^{-3}\}$ and number of layers $\in \{5, 10, 15\}$. The model was trained for 150 epochs with no learning rate schedule. The final configuration used \texttt{hidden\_dim} = 256, \texttt{n\_layers} = 10, and a learning rate of $10^{-4}$. For branch perturbation experiments, the best $\sigma$ on IEEE-118 was $\sigma = 0.01$.

\end{itemize}

Table~\ref{tab:capacity} reports the number of trainable parameters of each tuned model.

\begin{table}[h]
\caption{Number of trainable parameters of each tuned model.}
\label{tab:capacity}
\centering
\begin{tabular}{lr}
\toprule
\textbf{Model} & \textbf{Trainable parameters} \\
\midrule
CANOS-PF     & 9{,}560{,}710 \\
PowerFlowNet & 1{,}391{,}884 \\
GraphConv    & 1{,}057{,}030 \\
\bottomrule
\end{tabular}
\end{table}

\newpage
\section{Worst-case per-bus mismatch}
\label{appendix-worstcase}

\begin{figure}[!ht]
  \centering
  \includegraphics[width=\linewidth]{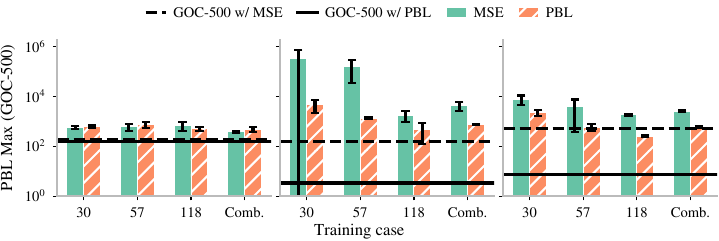}
  \\
  \begin{subfigure}[t]{0.32\linewidth}\caption{CANOS-PF}
  \end{subfigure}
  \begin{subfigure}[t]{0.32\linewidth}\caption{PowerFlowNet}
  \end{subfigure}
  \begin{subfigure}[t]{0.32\linewidth}\caption{GraphConv}
  \end{subfigure}
  \caption{Effect of loss function and multi-grid training on worst-case GOC-500 performance
           across all three architectures. Horizontal lines are as defined in Figure~\ref{fig:multigrid-all}.}
    \label{fig:multigrid-max-all}
\end{figure}

\begin{figure}[!ht]
    \centering
    \includegraphics[width=\linewidth]{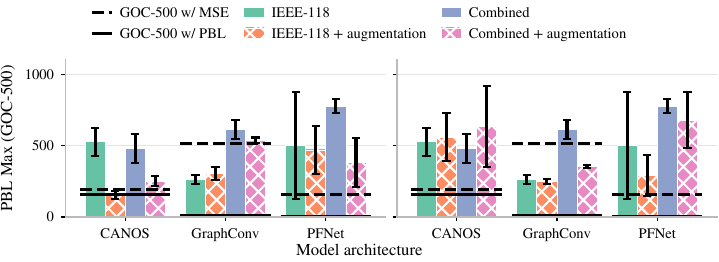}
    \\[-0.3em]
    \begin{subfigure}[t]{0.49\linewidth}
        \centering
        \subcaption{Branch perturbation}
        \label{fig:aug_max_branch_perturbation}
    \end{subfigure}
    \hfill
    \begin{subfigure}[t]{0.49\linewidth}
        \centering
        \subcaption{Line/generator N-2 dropout}
    \end{subfigure}
    \caption{Effect of the two data augmentation strategies on worst-case GOC-500 performance under PBL training. Horizontal lines are as defined in Figure~\ref{fig:multigrid-all}.
    }
    \label{fig:augmentation-max-all}
\end{figure}

Mean PBL averages the complex power mismatch over all buses and can hide violations concentrated at a small number of them. We repeat the analysis of Section~\ref{sec:experiments} using the maximum per-bus mismatch $\max_{i \in \mathcal{B}} |\Delta S_i|$ on GOC-500 (Figures~\ref{fig:multigrid-max-all} and~\ref{fig:augmentation-max-all}).

The conclusions are unchanged. Physics-informed training reduces worst-case mismatch for GraphConv and PowerFlowNet, with CANOS-PF unchanged. Training on a combined set of grids continues to perform similarly to training on the best single grid (IEEE-118).

Branch perturbation reduces CANOS-PF's worst case by a factor of two to three on IEEE-118 and on the combined set, has no consistent effect on GraphConv, which does not consume edge attributes, and is mixed for PowerFlowNet. Finally, line/generator dropout shows no consistent effect.

In addition to these conclusions, we note a few other observations. First, aggregate means substantially understate transfer failure. In zero-shot settings, worst-case bus mismatch can exceed the mean by up to 45$\times$, reaching 775 where the mean is 17. The failure to generalize to out-of-distribution settings is clearly visible at the aggregate, but is even more pronounced at the tail.

Second, the best zero-shot configuration for CANOS-PF (Figure~\ref{fig:aug_max_branch_perturbation}) attains a worst-case mismatch that is almost equivalent to that obtained by training directly on GOC-500. This is not replicated across the other models, suggesting CANOS-PF's edge-aware architecture responds particularly well to branch perturbation.

\newpage
\section*{AI4PowerGrids evaluation checklist}

\paragraph{Physics feasibility.}
\textit{For methods producing grid quantities: is feasibility reported under the full AC power-flow equations, not only the linearised DC approximation? Include constraint-violation statistics, not just aggregate error.}

Addressed. All results are reported under the full AC power flow equations. Our evaluation metric is mean Power Balance Loss, the per-bus complex power mismatch $|\Delta S_i|$ of the model's own predictions; no DC linearization is used at any point in training or evaluation. Appendix~\ref{appendix-worstcase} additionally reports the maximum per-bus mismatch, so violations concentrated at a few buses are not hidden by the average.

\paragraph{Out-of-distribution evaluation.} \textit{Is the method evaluated on at least one network topology, generation mix, or operating regime not seen in training?}

Addressed. We focus our results on zero-shot transfer to GOC-500, a 500-bus system absent from training (IEEE-30/57/118), so both topology and scale are unseen. Training and test sets additionally span $N-1$ and $N-2$ perturbations.

\paragraph{Failure modes.} \textit{Are failure cases reported alongside aggregate metrics – worst-case behaviour, tail statistics, or qualitative examples?}

Addressed. We report variability across training runs (error bars in Figures~\ref{fig:multigrid-all}, ~\ref{fig:augmentation-all}, ~\ref{fig:multigrid-max-all}, and \ref{fig:augmentation-max-all}) and report negative results (such as multi-grid training failing to close the transfer gap and line/generator dropout degrading performance) rather than only the interventions that worked. Finally, we characterize worst-case behavior in Appendix~\ref{appendix-worstcase}.

\paragraph{System-level effect (where applicable).} \textit{For components inside a larger system, is the downstream effect reported, not only standalone accuracy?}

Not applicable. We evaluate power flow approximation as a standalone task. Downstream effects within contingency screening or ACOPF pipelines are out of scope for this study.

\paragraph{Data and code.} \textit{Will code and data be made available (anonymously at submission, or on acceptance)? A clear release plan is sufficient. Proprietary code or restricted datasets are acceptable: state clearly what cannot be released and why, and describe what can be shared (aggregate metrics, synthetic surrogates, algorithm details).}

Addressed. All experiments use the publicly available PF$\Delta$ benchmark dataset \citep{rivera2026pf}. Code, trained configurations, and reproduction scripts will be released at \url{https://github.com/MOSSLab-MIT/pfdelta}.

\end{document}